\documentclass{article}

    \usepackage[nonatbib, preprint]{neurips_2019}

\usepackage[utf8]{inputenc} 
\usepackage[T1]{fontenc}    
\usepackage{hyperref}       
\usepackage{url}            
\usepackage{booktabs}       
\usepackage{amsfonts}       
\usepackage{nicefrac}       
\usepackage{microtype}      
\usepackage{graphicx}
\usepackage{subcaption}
\usepackage{geometry}
\usepackage{multirow}
\newcommand{\etal}{\textit{et al.}}

\title{``You might also like this model'':\\ Data Driven Approach for Recommending Deep Learning Models for Unknown Image Datasets}

\author{%
  Ameya Prabhu
  \And
  Riddhiman Dasgupta\\
  Microsoft AI  \\
  Hyderabad, India 
  \And
  Anush Sankaran, Srikanth T, Senthil Mani\\
  IBM Research AI\\
  Bengaluru, India
}

\begin{document}

\maketitle

\begin{abstract}
For an unknown (new) classification dataset, choosing an appropriate deep learning architecture is often a recursive, time-taking, and laborious process. In this research, we propose a novel technique to recommend a suitable architecture from a repository of known models. Further, we predict the performance accuracy of the recommended architecture on the given unknown dataset, without the need for training the model. We propose a model encoder approach to learn a fixed length representation of deep learning architectures along with its hyperparameters, in an unsupervised fashion. We manually curate a repository of image datasets with corresponding known deep learning models and show that the predicted accuracy is a good estimator of the actual accuracy. We discuss the implications of the proposed approach for three benchmark images datasets and also the challenges in using the approach for text modality. To further increase the reproducibility of the proposed approach, the entire implementation is made publicly available along with the trained models.
\end{abstract}

\graphicspath{{./images/}}
\section{Introduction}



With the current unprecedented growth in deep learning, the primary and most pressing challenge faced by the community is to find the most appropriate model for a given dataset. Consider the \$1M Data Science Bowl challenge for detecting lung cancer hosted by Kaggle in $2017$\footnote{\url{https://www.kaggle.com/c/data-science-bowl-2017}}. It introduces a dataset of lung scans, consisting of thousands of images, and aims to develop algorithms that accurately determine when lesions in the lungs are cancerous. To solve this in practice, a common approach is to abstract the problem of lung cancer detection as a special case of object detection: use pre-trained models on large scale image classification datasets, and fine-tune them to target lung cancer dataset.
The process would begin with choosing a state-of-the-art deep learning architecture, say \textit{AlexNet}~\cite{krizhevsky2012imagenet}, with weights pre-trained on ImageNet dataset. After fine-tuning multiple models and using multiple pre-training datasets, it is found that both the AlexNet architecture and ImageNet dataset are not suitable for the task of lung cancer detection. This procedure is then extensively repeated for multiple models, such as \textit{ResNet}~\cite{he2016deep}, \textit{VGG-16}~\cite{simonyan2014very}, \textit{VGG-19}~\cite{simonyan2014very}, and \textit{Network-in-Network}~\cite{lin2013network}, till the ideal model is found. Similarly, it has to be repeated for different datasets such as CIFAR-10, CIFAR-100, and TinyImageNet until the pre-training dataset is obtained. This is an extremely expensive hit-and-trial search approach.

Models pre-trained from generic datasets have shown improved performance in different domains and diverse tasks such as music genre classification~\cite{raina2007self}, face recognition~\cite{sun2014deep}, healthcare~\cite{esteva2017dermatologist}, and food industry~\cite{wang2015recipe}. Currently, the choice of the generic dataset and pre-train model is purely based on human expertise and prior knowledge. Kornblith et al.~\cite{kornblith2018better} considered thirteen different deep learning models trained on ImageNet and studied the fine-tuned performance to different target datasets. They found improvements in certain transfer scenarios and also showed that transferability is limited based on the source and target datasets properties. This explains the necessity for a systematic approach to choose the dataset and pre-trained networks for a given unknown dataset or task (such as, lung cancer detection). In this paper, we aim to address this problem by proposing an automated deep learning model recommendation system from a repository of models for a given unknown dataset. We further predict the accuracy of the recommended deep learning model on the unknown dataset without the need for training or fine-tuning. This enables the user to take a well informed decision on which popular deep learning model to adopt for the unknown dataset in hand and also what is the ballpark performance to expect. 

The proposed research problem is defined as follow: For a given unknown dataset $d_u$, select a dataset $d_c$ and model $m_c$ that will provide the best fine-tuning accuracy, $a_c$ of $d_u$ after being pre-trained on $d_c$ and also predict the accuracy $a_c$ without the need for training. Formally, our system assumes a repository of $k$ popular deep learning architectures trained independently on $n$ different existing datasets. Given an unknown dataset $d_u$, we find the most similarly dataset $d_c$ from the list of $n$ datasets and predict the accuracy of every model $k$ on that dataset, without actually training the model on it. This allow us to quantitatively assess the promise of transferring models and also recommend a suitable model for the unknown dataset.
For example, for the given unknown lung cancer dataset, say the proposed approach predicts that STL-10 dataset~\cite{coates2011analysis} as the most similar dataset. Then, we predict the accuracy of all the architectures available for STL-10 in the model repository for the lung cancer dataset and rank them. Thus, we obtain the best performing pre-training dataset as well as the architecture. The proposed approach advances the literature to achieve a deep neural network recommendation systems using only limited resources and in real-time.
To summarize, the primary research contributions of this research are as follows:

\noindent1. \textit{A model recommendation system} which predicts the best suitable pre-trained model from a repository of models and predict its accuracy for the unknown dataset.

\noindent2. A general purpose unsupervised \textit{model encoder} which extracts a fixed length, continuous vector representation for any given discrete, variable-length deep learning architecture, along with its hyperparameters.

\noindent3. A dataset \textit{similarity ranker} system which characterizes the similarity distribution between a given unknown dataset and datasets in our repository using an ensemble of classifiers. We show that it is possible to get a good correlation between the dataset similarity predictions and actual accuracy obtained on that dataset.

\noindent4. A \textit{accuracy regressor} which estimates the accuracy of a deep learning model on an unknown query dataset efficiently, using the dataset similarity ranker and model encoder features.

\noindent5. In order to further increase the reproducibility of the proposed work, the entire working implementation is publicly made available along with the trained models: \url{https://github.com/dl-model-recommend/cikm-2019}


\section{Existing Literature}
We will discuss the existing literature in the area of Neural Architecture Search (NAS), accuracy prediction, as well as recommender systems.

\noindent \textbf{Neural Architecture Search (NAS):} 
The aim of NAS is to find the most suitable architecture for a given dataset from the set of all possible architectures~\cite{zoph2017neural}.
ENAS~\cite{pham2018efficient} is the first work towards fast and inexpensive automatic model design. 
Baker~\etal~\cite{baker2017designing} uses a Markov decision based meta-modeling algorithm with an average time to search for the best model being around 8-10 days. 
Liu~\etal~\cite{liu2018hierarchical} and Real~\etal~\cite{real2018regularized} use an evolutionary algorithm instead of reinforcement learning algorithms. 
Liu~\etal~\cite{liu2017progressive} propose using a sequential model-based optimization (SMBO) strategy, which is up to $5-8$ times more efficient than reinforcement learning based techniques.
Liu~\etal~\cite{liu2018darts} is the first major work to pose architecture search as a differentiable problem over a discrete and non-differentiable search space instead of a reinforcement learning problem. 

\noindent \textbf{Accuracy Prediction:} 
Baker~\etal~\cite{baker2018accelerating} leverage standard frequentist regression models to predict final performance based on architecture, hyperparameters and partial learning curves. 
Deng et al.~\cite{deng2017peephole} predict the performance of a network before training, based on its architecture and hyperparameters. 
TAPAS~\cite{istrate2018tapas} is another novel deep neural network accuracy predictor, parameterized on network topology as well as a measure of dataset difficulty. 
Scheidegger~\etal~\cite{scheidegger2018efficient}  introduced a class of neural networks called ProbeNets to measure the difficulty of an image classification dataset, without having to train expensive state-of-the-art neural networks on new datasets. 
In contrast to existing techniques that rely on reinforcement learning or evolutionary algorithms, Elsken~\etal~\cite{elsken2018simple} employ a new method which is a combination of hill climbing, network morphism, and cosine annealing based optimization. Summarizing, works such as Peephole~\cite{deng2017peephole} use only the model architecture, while TAPAS~\cite{istrate2018tapas} use the model architecture along with the characterization of the query unknown dataset. In the proposed work, we use the model architecture as well as the similarity between the unknown dataset and the known dataset on which the model was trained upon. Additionally, the training process for learning the model representation and the dataset similarity are performed with a large training data.

Traditionally in literature, deep learning methods are used as a solution to solve the personalized recommendation problem. However in this research, we propose a technique to use recommendation systems as a solution for which deep learning model to be used for a dataset and task. Further, most of the existing NAS techniques for deep learning are still unusable in practical situations, requiring huge clusters of GPUs and consuming a lot of time\footnote{\url{https://twitter.com/beenwrekt/status/961262527240921088}}. Moreover, in most of these applications, finding a novel architecture from scratch is not essentially required and a minor variant of a popular deep learning model would suffice.

\begin{figure*}[t!p]
    \begin{subfigure}[b]{0.48\textwidth}
        \centering
        \includegraphics[width=0.98\textwidth]{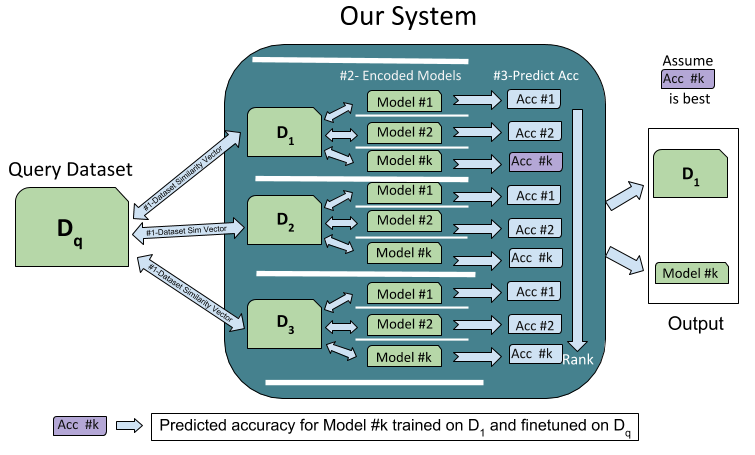}
        \caption{An overview of the proposed system.}
        \label{fig:overview2}
    \end{subfigure}
    \hfill
    \begin{subfigure}[b]{0.48\textwidth}
        \centering
        \includegraphics[width=0.98\textwidth]{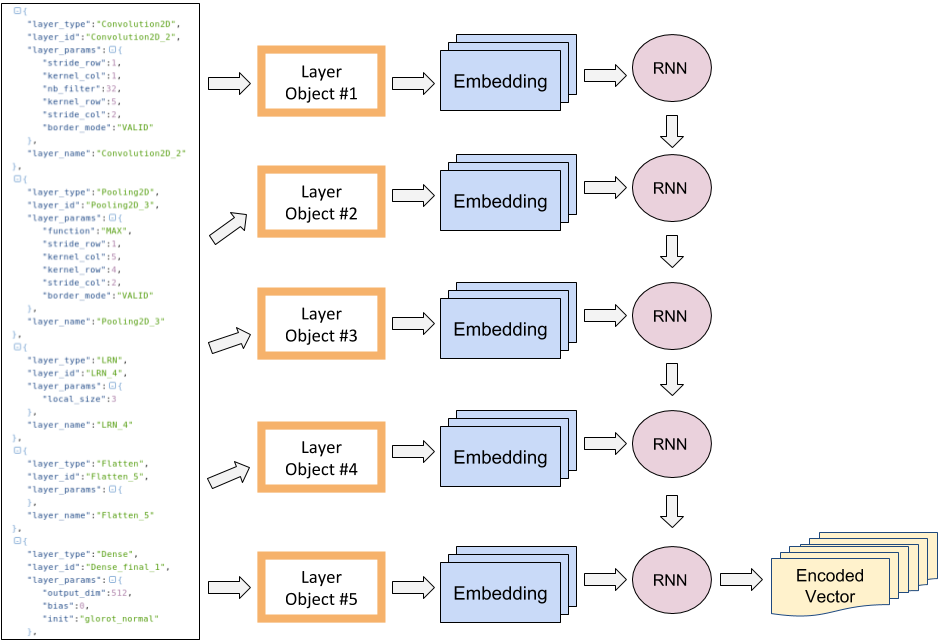}
        \caption{An overview of the Unsupervised Model Encoder}
        \label{fig:dataset2vec2}
    \end{subfigure}
\caption{Given a query dataset, we first calculate the dataset similarity vector. The obtained pairwise vector along with the model encoding is used to predict the accuracy. Then we rank the results and recommend a model from our repository.} \label{newfig1}
\end{figure*}

\section{Model Recommendation Approach}

As illustrated in Figure~\ref{fig:overview2}, the proposed approach consists of three novel components: 
\begin{enumerate}
    \item \textbf{Unsupervised Model Encoder:} which obtains a fixed length continuous space representation for a variable-length, discrete-spaced deep learning model architecture, along with its hyperparameters, using an unsupervised encoding technique.
	\item \textbf{Dataset Similarity Ranker:} which predicts the most similar existing dataset $d_c \in [d_1, d_2, d_3, \ldots, d_n]$ for any given unknown dataset $d_u$.
	\item \textbf{Accuracy Regressor:} It learns the mapping from the above two unsupervised representations to the accuracy obtained by the model.
\end{enumerate}
Thus, for a given unknown dataset, our system will retrieve a dataset and architecture from the repository using the dataset similarity ranker, encode a fixed length representation of the architecture using the unsupervised model encoder, and predict the accuracy of the architecture on the unknown dataset using the accuracy regressor. Although this is quite a challenging combination of tasks, we feel that it remains an important problem to solve, given its benefits in saving both resources and time compared to hit-and-trial approaches.

\subsection{Unsupervised Model Encoder}
Deep neural networks' architecture can be considered as a directed acyclic graph (DAG) whose nodes represent certain transformations, such as convolution, recurrent cells, dropout, and pooling. In this component, we aim to develop a representation of such a graph (network architecture) in an unsupervised fashion. The first step is to define a representation of individual nodes, i.e., the layers and encode information about the layer sequence into fixed sized vectors. This is analogous to encoding individual words of a sentence using a word embedding model (such as, \textit{word2vec}) and using the individual word embeddings to learn a language model at the sentence level.

\textbf{Learning to generate valid models:} We exploit the fact that models have only certain structures which are \textit{valid}. \textit{Valid} models are those which could be trained for a given dataset without any errors and could turn out to be \textit{accurate / optimal} or \textit{inaccurate / sub-optimal} for that dataset. \textit{Invalid} models are those that are either structurally impossible to occur, such as networks having embedding layer between two LSTM layers, or those that cannot be compiled for the given dataset, such as a CNN that reduces the image size to less than zero. Similarities can be drawn between this imposition of structures in deep networks and imposition of a grammar in a language. This further motivates the usage of a sequential language model technique to encode possible structures of a network architecture. A manually defined grammar is used to generate lots of possible valid models for a given dataset and these valid models are stored in a custom JSON structure, which is very similar to the Keras JSON format or the Caffe protobuf format.

\textbf{Construction:}
As illustrated in Figure~\ref{fig:dataset2vec2}, given an input abstract JSON representation of model architecture, we compute a fixed-length vector as the output. The major steps are as follows:\\ 
\noindent (1) Layer Encoding: A layer vocabulary is constructed which contains all unique layers with its hyperparameter combinations. For instance, a Convolution2D layer has the following hyperparameter set: \{'number of filters': [512, 384, 256, 128, 64, 32], 'kernel row': [1, 2, 3, 4, 5], 'kernel column': [1, 2, 3, 4, 5], 'stride row': [1, 2, 3], 'stride column': [1, 2, 3], 'border mode': ['Same', 'Valid']\}, totalling to $2700$ unique combinations to the layer vocabulary. To account for layers or hyperparameters that are not a part of our grammar, we added an \textit{Unknown} layer, UNK, to our vocabulary to be able to encode any kind of deep learning architecture. A total of $19$ unique layers were used resulting in a vocabulary size of $4523$ tokens.  The encoding is performed similar to a Unified Layer Code\cite{deng2017peephole}. \\
\noindent (2) Generating Layer Representations: Each model architecture is represented as a sequence of tokens, for example \textit{Convolution2D \_512\_3\_3\_1\_1\_Same} is one token in that sentence. If a function model is provided, each path from source to sink is added as an independent sentence. Inspired from word embedding, for each given layer we predict the surrounding context of layers resulting in vector representations for each layer, independently. We train \textit{word2vec} representation with standard hyperparameters (\textit{gensim} library) to obtain a $512$-dimensional layer representation. \\
\noindent (3) Generating Model Representations: We use the layer embeddings to initialize and train a three layer LSTM model with tied weights and trained it similar to a language modelling task to generate the 512-dimensional model representation. Sentence perplexity is used as the objective function to be optimized while learning the language model.

Thus, we develop an unsupervised subsystem to  convert a variable length sequence of discrete network layers to a succinct, continuous, vector-space representation.

\begin{figure*}[t!p]
    \begin{subfigure}[b]{0.48\textwidth}
        \centering
        \includegraphics[width=0.98\textwidth]{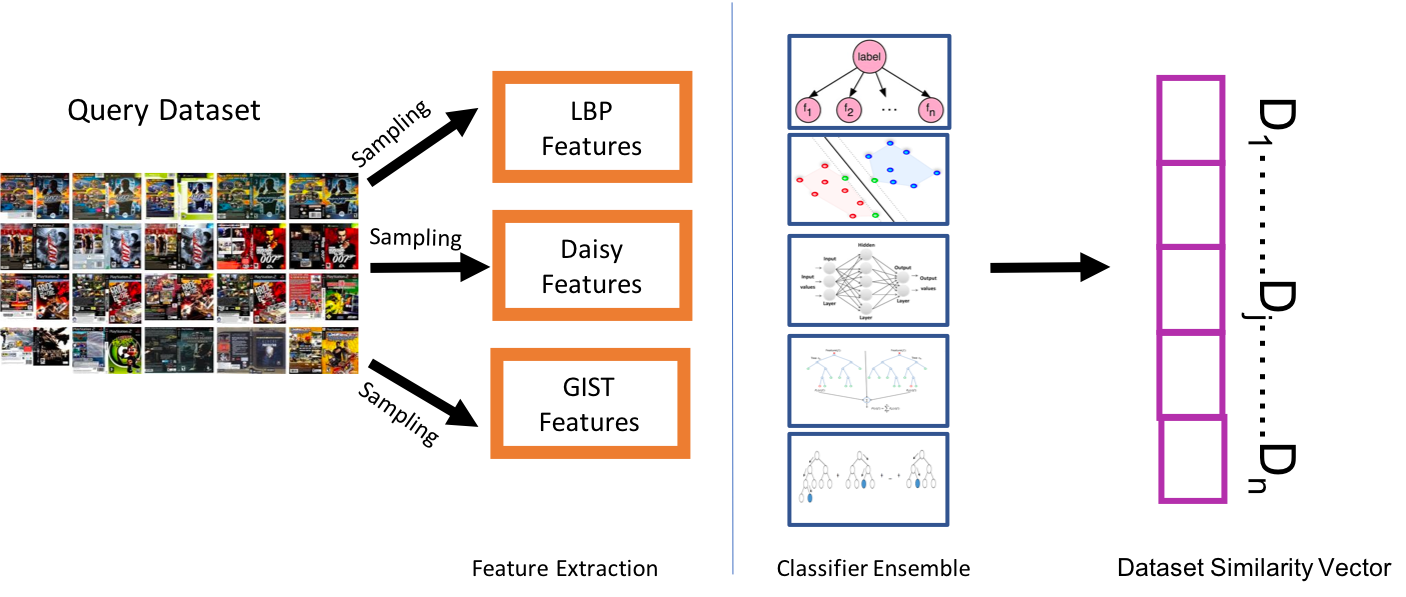}
        \caption{An overview of the Dataset Similarity  Ranker.}
        \label{fig:dataset2vec1}
    \end{subfigure}
    \hfill
    \begin{subfigure}[b]{0.48\textwidth}
        \centering
        \includegraphics[width=0.98\textwidth]{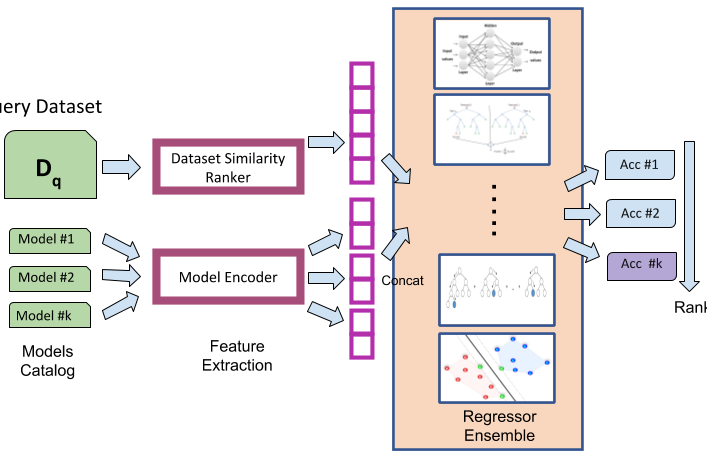}
        \caption{An overview of the Accuracy Regressor}
        \label{fig:dataset2vec3}
    \end{subfigure}
\caption{Given the dataset and model encoding representations, we can compute the predicted accuracies for that pre-trained pairing. In this manner, we predict the accuracies. They are further ranked and the best predicted accuracy is used to return the model and dataset.} \label{newfig2}
\end{figure*}

\subsection{Dataset Similarity Ranker}
This component computes the similarity between the given dataset and all existing datasets in the dataset repository. The aim is to study the similarity between datasets and provide a guided approach for transferability between datasets.
As illustrated in Figure~\ref{fig:dataset2vec1}, given a query dataset, $d_q$ and a list of repository datasets, $d_i, i \in [1,\ldots, n]$, the procedure for calculating the dataset similarity between the query dataset and the repository datasets is as follows:

1. A set of $s$ data samples are uniformly picked from each of the repository datasets $d_i$.

2. For every sample, $j$, in these $d_i$, we extract features $f_{ij}$ from the input data. These form the input vectors and the output class is the dataset number $i$.  

3. Several classifiers are trained on each of the sampled image features to predict which of the $n$ repository datasets does the given feature vector belong to. Torralba~\cite{torralba2011unbiased} studied the presence of a unique signature for every dataset, enabling us to find similarity and dissimilarity between datasets.

Now, given an unknown query dataset, $d_u$

1. A set of $s$ data samples are randomly picked from the query dataset, $d_u$. For each sample, we extract all the set of features $f_u$.

2. The features are passed to the respective trained $n$-class classifiers, which classify each sample individually to one of the $n$ repository datasets,
	\begin{equation}
    	\sum_{j=1}^{ns} C_{k,f}(s^{(j)}_u) 
	\end{equation}
	$\forall  \{k,f\} \in [1,\ldots, en]$, $C_{k,f}$ denotes the classifier $k$ learnt on feature $f$, and $ns$ is the number of samples in the set $s$

3. We collect all the predictions and perform majority voting fusion across the ensemble, obtained a $n\times 1$ output vector denoting the probability of the similarity between the unknown dataset $d_u$ against each of the $n$ repository dataset $d_i$,
	\begin{equation}
	\oplus_{i} sim(d_q,d_i) = \sum_k \sum_f \sum_{j=1}^{ns} C_{k,f}(s^{(j)}_u)
	\end{equation}
where, $\oplus_{i}$ denotes concatenation of values across the $i$ repository datasets

There are three feature extractors used for the image modality: (i) GIST~\cite{oliva2006building} (ii) DAISY~\cite{tola2010daisy} (iii) Local Binary Pattern (LBP)~\cite{zhang2007face}. Five popular classifiers are used in the ensemble: (i) Naive Bayes (NB), (ii) Random Decision Forest (RDF), (iii) Boosted Gradient Trees (BGT), (iv) Multilayer Perceptron (MLP), and (v) Support Vector Machines (SVM). 

\subsection{Accuracy Regressor}
The accuracy regressor takes a $n_1$-dimensional dataset similarity vector between unknown dataset $d_u$ and repository dataset $d_i$ (obtained using equation (1) and a $n_2$-dimensional model representation vector as input and predicts accuracy of the model for the unknown dataset, as shown in Figure~\ref{fig:dataset2vec3}. This is learnt using a supervised regression approach, thus avoiding the need to efficiently learn to predict accuracy of deep networks. The system predicts the expected accuracy of a model trained on a dataset with a degree of similarity to the query dataset as given by the dataset similarity vector. 

Given a query dataset $d_u$, the accuracy regressor component is learnt as follows: 

1. We extract the dataset similarity vector, $n_1$, for every pair of dataset (all seven image datasets) using the dataset similarity ranker subsystem. 

2. Using model encoder subsystem, we encode the models available in our model repository to obtain a vector $n_2$ for each model.

3. We concatenate these two features as $n_1 + n_2$ dimensional input vector and perform regression using an ensemble of regressors to learn a mapping function between this high dimensional input vectors and the accuracy of the model, pre-trained on $d_i$ and fine-tuned on $d_u$.

We use eight different types of regressors: (i) Support Vector Regressor (RBF, linear, polynomial Kernel), (ii) Multi-Layer Perceptron, (iii) Ridge Regression, (iv) RandomForest Regressor, (v) GradientBoosting Regression, and (vi) AdaBoost Regressor. 


\section{Experiments and Analysis}
In this section, we demonstrate the performance of the three individual components and the overall approach. All the experiments were implemented using PyTorch~\footnote{\url{https://pytorch.org/}} and the code is publicly made available along with the trained models: \url{https://github.com/dl-model-recommend/cikm-2019}

\subsection{Model Repository}
The image dataset repository contains seven different diverse benchmark vision datasets: (i) MNIST, (ii) Fashion-MNIST, (iii) CIFAR-10, (iv) CIFAR-100, (v) SVHN, (vi) STL-10, and (vii) GTSRB. All of them are resized to $32 \times 32$ pixels. 
The choice of image based deep learning architectures in the repository is constrained by the input image size ($32\times 32$), with: (i) VGG-16~\cite{simonyan2014very}, (ii) Network-in-Network (NIN)~\cite{lin2013network}, (iii) Strictly Convolutional Neural Network (All-CNN)~\cite{springenberg2014striving}, (iv) ResNet-20~\cite{he2016deep}, (v) Wide-ResNet~\cite{zagoruyko2016wide}, (vi) Pre-ResNet~\cite{he2016identity}, and (vii) LeNet~\cite{lecun1998gradient}. 

\subsection{Experimental Details}
To learn the word embedding and the language model for unsupervised model encoding, we generated $190,000$ random valid models using the proposed grammar (simulated dataset). For each model, we randomly replaced a layer as \textit{UNK} with a probability of $0.2$ and generated a total of $570,000$. This dropout makes the sampling more diverse as well as enables us to encode models which cannot be defined by the grammar.
To train and evaluate the accuracy regressor, we take a subset of models from the above set of $190,000$ models. We train these models on the seven different image datasets and we have $700$ inaccurate models which perform poorly and $504$ accurate models on the respective datasets. This constitutes a total of 1204 models along with the accuracy they obtain on the respective datasets. We divide the models into a 80-20 train-test split randomly and use this dataset to train and evaluate the accuracy regressor. 


\begin{figure*}[t!p]
    \begin{subfigure}[b]{0.48\textwidth}
        \centering
        \includegraphics[width=0.98\textwidth]{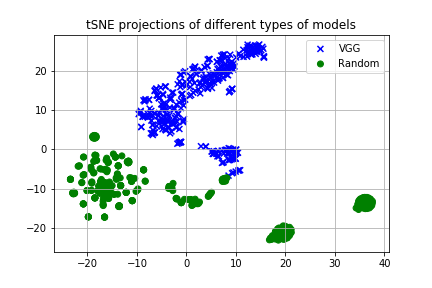}
        \caption{The tSNE plot of $70$ VGG variant and random DL models}
        \label{fig:dataset2vec56}
    \end{subfigure}
    \hfill
    \begin{subfigure}[b]{0.48\textwidth}
        \centering
        \includegraphics[width=0.98\textwidth]{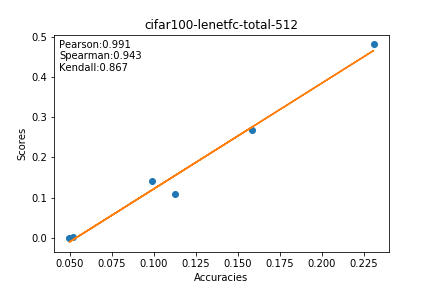}
        \caption{Correlation plot with coefficients for an unknown dataset}
        \label{fig:dataset2vec2}
    \end{subfigure}
\caption{The performance of unsupervised feature encoder} \label{newfig4}
\end{figure*}

\subsection{Unsupervised Model Encoder}
We evaluate the subsystem by evaluating the perplexity of the encoded representations generated, as shown in Figure~\ref{newfig4} (b). A lower perplexity score implies that the language model is better at generating valid models. To study the effectiveness of our learned model architecture representation, we take $70$ variations of VGG model by varying the number of blocks with hyperparameters and $70$ random deep learning models. The two dimensional tSNE visualization of the model representations in Figure~\ref{fig:dataset2vec56} show that all the VGG-like models are clustered together and are very different from the random deep learning models. This shows that similar looking architectures have similar representations in the learnt feature space. Thus, the proposed unsupervised model encoder can be used as a general purpose deep learning architecture encoding technique and can be used and extended for multiple applications.

\begin{figure*}[t!p]
    \begin{subfigure}[b]{0.48\textwidth}
        \centering
        \includegraphics[width=0.98\textwidth]{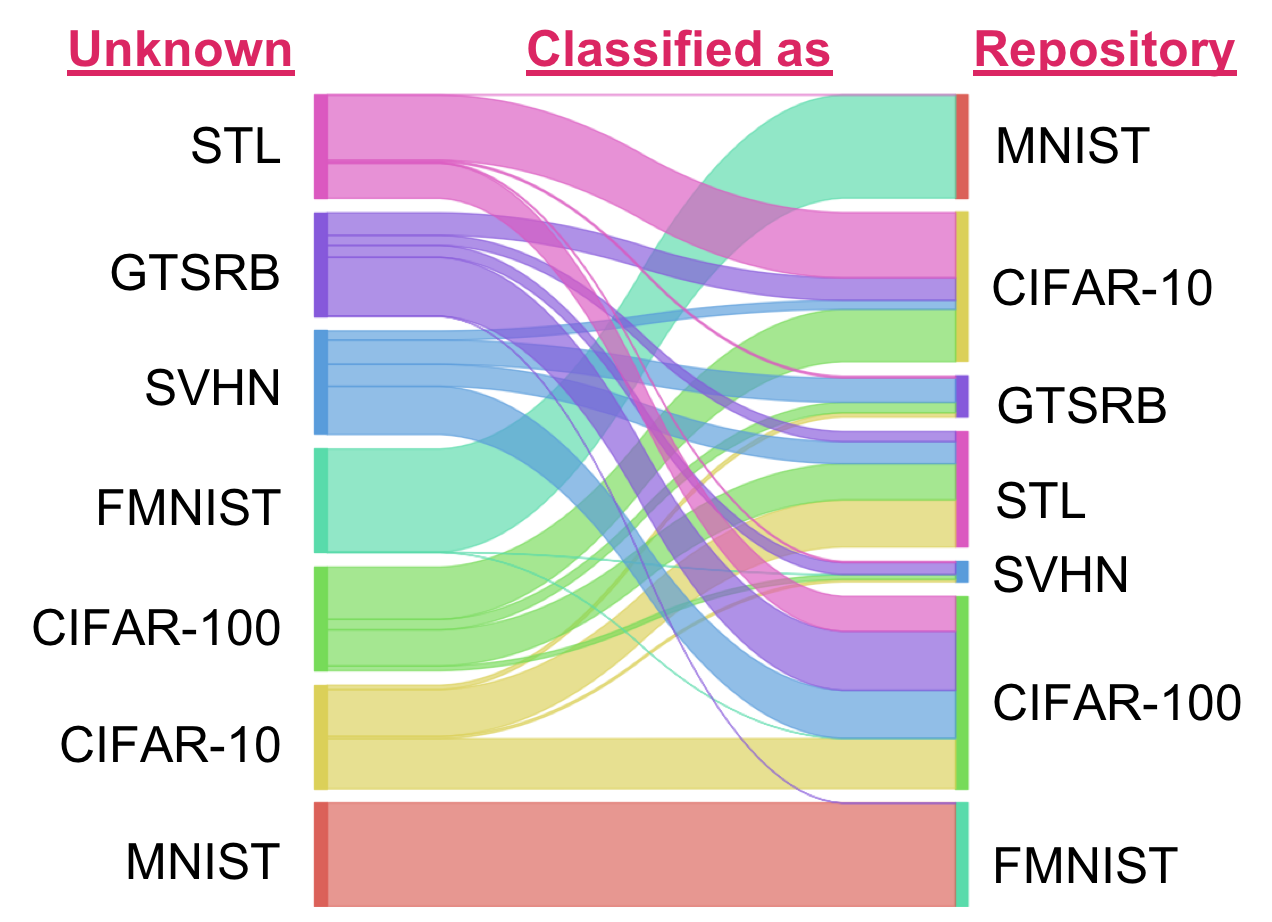}
        \caption{Sankey plot with computer dataset similarity}
        \label{fig:dataset2vec6}
    \end{subfigure}
    \hfill
    \begin{subfigure}[b]{0.48\textwidth}
        \centering
        \includegraphics[width=0.98\textwidth]{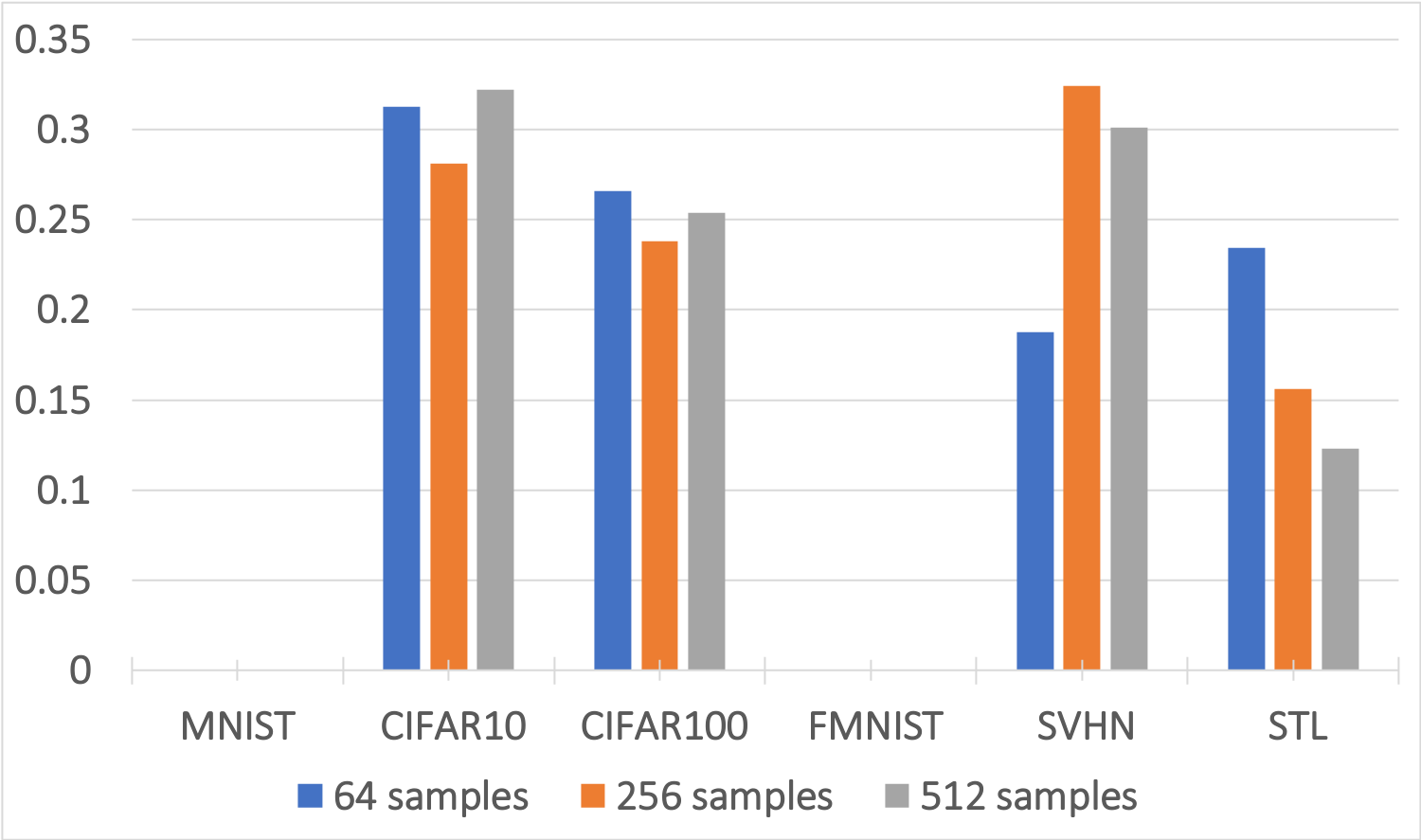}
        \caption{The effect of sample size}
        \label{fig:sample}
    \end{subfigure}
\caption{The performance of dataset similarity ranker} \label{newfig4}
\end{figure*}

\begin{table*}[!h]
	\centering
	\begin{tabular}{|c|c|c|c|c|c|c|}
		\hline 
		Unknown dataset  & \textbf{CIFAR10} & \textbf{CIFAR100} & \textbf{FMNIST} & \textbf{GTSRB} & \textbf{MNIST}  & \textbf{SVHN} \\ \hline
		\multicolumn{7}{c}{\textbf{VGG-16}}              \\ \hline
		\textbf{Pearson}  & 0.981            & 0.844             & 0.564           & 0.796          & 0.572                 & 0.217         \\ \hline
		\textbf{Spearman} & \textbf{0.928}            & \textbf{0.943}             & 0.883           & 0.886          & 0.429      & 0.486         \\ \hline
		\textbf{Kendall}  & 0.828            & 0.867             & 0.788           & 0.733          & 0.200                 & 0.333         \\ \hline
		
		\multicolumn{7}{c}{\textbf{NIN}}              \\ \hline
		\textbf{Pearson}  & 0.624 & 0.070 & 0.976 & 0.312 & 0.970   & 0.899  \\ \hline
		\textbf{Spearman} & 0.828 & 0.029 & 0.551 & 0.232 & 0.599    & 0.714       \\ \hline
		\textbf{Kendall} & 0.733 & 0.066 & 0.414 & 0.138 & 0.466   & 0.466  \\ \hline
		
		\multicolumn{7}{c}{\textbf{All-CNN}}              \\ \hline
		\textbf{Pearson}  & 0.480 & 0.492 & 0.983 & 0.085 & 0.978   & 0.934  \\ \hline
		\textbf{Spearman} & 0.314 & 0.486 & 0.464 & 0.232 & 0.314   & 0.714       \\ \hline
		\textbf{Kendall} & 0.200 & 0.333 & 0.276 & 0.138 & 0.200  & 0.466  \\ \hline
		
		\multicolumn{7}{c}{\textbf{ResNet-20}}           \\ \hline
		\textbf{Pearson}  & \textbf{0.95}             & 0.886             & 0.728           & 0.790          & 0.720                    & -0.185        \\ \hline
		\textbf{Spearman} & 0.638            & \textbf{0.943}             & 0.706           & 0.829          & 0.486                    & -0.029        \\ \hline
		\textbf{Kendall}  & 0.552            & 0.867             & 0.645           & 0.733          & 0.333                   & -0.067        \\ \hline
		
		\multicolumn{7}{c}{\textbf{Wide-ResNet}}              \\ \hline
		\textbf{Pearson}  & 0.000 & 0.089 & 0.966 & 0.499 & 0.969   & 0.697  \\ \hline
		\textbf{Spearman} & -0.085 & 0.257 & 0.522 & 0.232 & 0.486     & 0.609       \\ \hline
		\textbf{Kendall} & -0.200 & 0.200 & 0.414 & 0.138 & 0.333  & 0.414  \\ \hline
		
		\multicolumn{7}{c}{\textbf{Pre-ResNet}}              \\ \hline
		\textbf{Pearson}  & -0.370 & 0.721 & 0.962 & 0.497 & 0.965   & 0.945  \\ \hline
		\textbf{Spearman} & -0.085 & 0.428 & 0.521 & 0.232 & 0.486   & 0.714       \\ \hline
		\textbf{Kendall} & -0.200 & 0.200 & 0.414 & 0.138 & 0.333  & 0.466  \\ \hline
		
		\multicolumn{7}{c}{\textbf{LeNet}}\\ \hline
		\textbf{Pearson}  & \textbf{0.981}            & \textbf{0.991}             & 0.756           & 0.798          & 0.755    & \textbf{0.982}         \\ \hline
		\textbf{Spearman} & \textbf{0.928}            & \textbf{0.943}             & 0.530           & 0.886          & 0.600    & \textbf{1.00}          \\ \hline
		\textbf{Kendall}  & 0.828            & 0.867             & 0.501           & 0.733          & 0.467        & \textbf{1.00} \\ \hline 
	\end{tabular}
	\caption{\label{table1} The correlation coefficients obtained between the dataset similarity scores and the actual performance accuracy. This shows that the dataset similarity score is an unbiased estimator of the model's accuracy.}
\end{table*}

\subsection{Dataset Similarity Ranker}
We evaluate the performance of the dataset similarity ranker by performing an exhaustive leave-one-out test on the dataset repository. For each of the seven unknown datasets and the rest of the repository, we predict the ranking of the datasets obtained from our system. To obtain the ground truth, we train all the seven models: (i) VGG-16, (ii) NIN, (iii) All-CNN, (iv) ResNet-20, (v) Wide-ResNet, (vi) Pre-ResNet, and (vii) LeNet on each of the 6 remaining datasets present in the catalog. Given the query dataset, we fine-tune these networks, giving accuracy values $a_1, a_2, .., a_6$. The ensemble of models are trained using a sample of images taken from the train dataset of the respective datasets, while the ensemble of models are tested using a sample of images taken from the test dataset. The covariance shift that exists between the train and test of the respective datasets could also influence the performance of the dataset similarity ranker.
We obtain the correlation scores and show that the dataset ranking provided by our system is highly correlated to the ranking obtained by the accuracy exhaustive fine-tuning. This indicates that models pre-trained on the dataset that we predicted to be the most similar dataset, provided the best performance accuracy after being fine-tuned on the unknown dataset. The results are populated in Table~\ref{table1} and the correlation plot for an unknown dataset, CIFAR-100 and LeNet as the model is provided in Figure~\ref{fig:dataset2vec2}. It can be observed that the correlation coefficients are positive and high for all the datasets except SVHN. This implies that finding similar datasets that could provide a good pre-training for models is possible and also shows that there are no similar datasets for SVHN in the repository, indicating that none of the pre-trained models are bound to produce high results in SVHN datasets.

Also, based on general intuition we expect CIFAR-10 and CIFAR-100, and MNIST and Fashion-MNIST to look visually similar. The proportion of each unknown dataset being classified to the repository dataset is shown in figure~\ref{fig:dataset2vec6}, which follows our intuitions.
Furthermore, we study the effect of sample size which is one of the critical hyper-parameter for computing the dataset similarity. Although we used $512$ as the effective sample size, we studied the effect of four different sample size on the classification performance: [$64$, $256$, $512$]. The result is shown in Figure~\ref{fig:sample} and can be observed that our subsystem can give reliable predictions irrespective of the size of the sample for MNIST and CIFAR variations. However, for SVHN and STL for which there are no related datasets, a smaller sample size tends to classify the input images towards SVHN.

\begin{figure*}[t!p]
    \begin{subfigure}[b]{0.48\textwidth}
        \centering
        \includegraphics[width=0.98\textwidth]{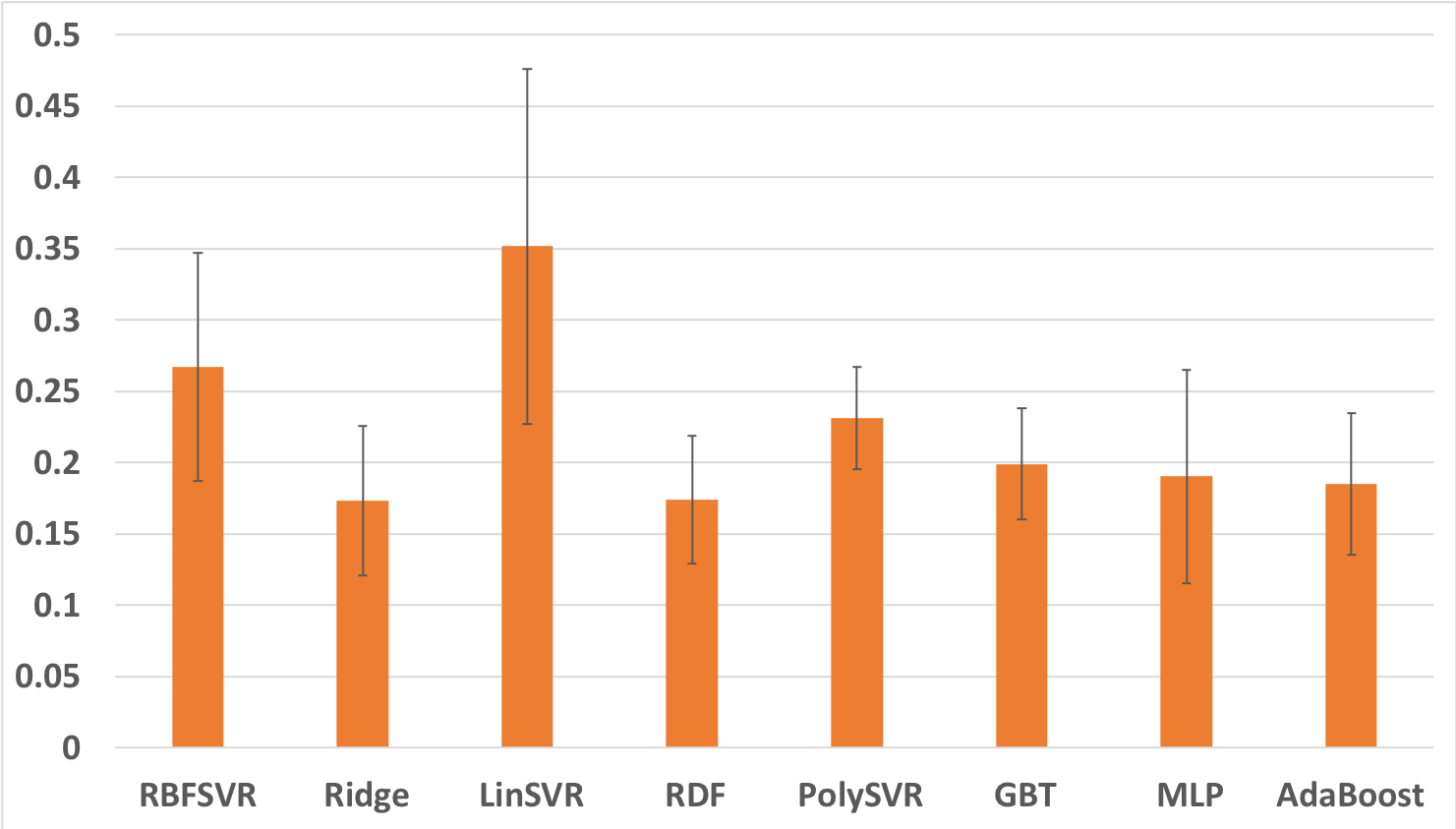}
        \caption{The MSE error of the various regressors}
        \label{fig:sample22}
    \end{subfigure}
    \hfill
    \begin{subfigure}[b]{0.48\textwidth}
        \centering
        \includegraphics[width=0.98\textwidth]{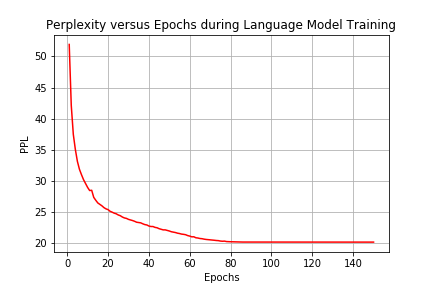}
        \caption{The perplexity graph of training an LSTM for Unsupervised Model Encoder.}
        \label{fig:sample}
    \end{subfigure}
\caption{The performance of accuracy regressors and the reason for failure in text based DL models} \label{newfig4}
\end{figure*}

\subsection{Accuracy Regressor}
We evaluate the regressor model using the Mean Square Error (MSE) error between the predicted accuracy by the regressor and the actual accuracy obtained after fine-tuning. The obtained results are shown in Figure~\ref{fig:sample22}. It can be observed that ridge regression performs the best with a MSE of $0.15$. This shows the a simple regression could predict the approximate performance of a deep learning model on a given unknown dataset, without the need for sophisticated models. 
Thus for a given unknown dataset, we sample $n=512$ images, find the most similar dataset using an ensemble of simple machine learning classifier. For all the architectures available in the repository for most similar dataset, we extract a fixed length representation using the unsupervised model encoder. This is a simple forward pass through the word embedding layer and the LSTM based language model. The dataset similarity vector and the model representation is fed into the accuracy regressor to predict the performance of the given models and find the best performing architecture. Hence, we show that accuracy prediction could be a practical almost real-time solution and could be adopted to various challenging domains.

\section{Practical Use Case}
A practitioner will usually prefer the most recent deep learning model, which might be unnecessarily complex for the task at hand. However, theoretically the choice of model depends on the properties of the dataset and the task~\cite{cnnbaseline}. It is interesting to study the performance of the proposed model recommendation system with respect to human preferences.
To show the effectiveness of the proposed deep learning model recommendation pipeline in a practical setting, we provide human baselines for three different datasets: (i) Caltech-UCSD Birds-200-2011~\cite{WelinderEtal2010}, (ii) Stanford Cars~\cite{krause20133d}, and (iii) ETHZ Food-101~\cite{bossard14}. Accuracies of various deep learning learning models on these datasets are manually computed in the literature~\cite{cnnbaseline}. For Caltech-UCSD Birds-200-2011 and ETHZ Food-101, our approach retrieved ResNet as the recommended architecture with a predicted accuracy of $63.2\%$ and $57.4\%$, respectively. The ground truth training, as performed in the literature~\cite{cnnbaseline}, yields $76.3\%$ and $67.59\%$, respectively, which are much higher than LeNet and VGG models. However, in case of Stanford Cars dataset, our approach recommended VGG-16 architecture with a predicted accuracy of $82.1\%$. This trend could be observed in the literature, as well, where VGG-16 performs better than ResNet variants and LeNet providing $85.2\%$ accuracy.
Thus, although the accuracy prediction provides a ballpark of the expected actual accuracy, the rank order of the retrieved models suggests that the proposed approach does not always retrieve the most complex model, but rather, retrieves models based on the properties of the datasets, the task, and the architecture of the model. 

\section{Conclusion and Future Work}
We proposed a novel system for recommending the most suitable pretrained architecture for a given unknown dataset. The proposed system consists of 3 subsystems: a dataset similarity subsystem, which predicts the similarity for any two given datasets; an unsupervised model encoder which  extracts a fixed length, continuous vector representation and an accuracy regressor which estimates the accuracy of a deep learning model on a unknown query dataset. Combining these subsystems, we explore the aim of recommending neural network models. Our system is one of the earliest approaches in this direction, and we hope that this research acts as a seed work for future extensions.

{\small
    \bibliographystyle{acm}
    \bibliography{egbib}
}

\newpage

\begin{center}
	\Large \textbf{Supplementary Material}
\end{center}

In order to further increase the reproducibility of the proposed work, the entire working implementation is publicly made available along with the trained models: \url{https://github.com/dl-model-recommend/cikm-2019}

\section{Additional Details: Dataset Similarity Ranker}
\noindent The model catalog has seven image datasets. The properties and details of each of these datasets are provided in Table 1.
\begin{table*}[!h]
\centering
\begin{tabular}{|l|l|l|l|l|}
    \hline
    Dataset                         & Image Size                        & \#Classes & TrainSize & TestSize  \\ \hline \hline
    \multirow{1}{*}{MNIST}          & $28\times 28$                     & $10$      & $50000$   & $10000$   \\ \hline
    \multirow{1}{*}{FashionMNIST}   & $28\times 28$                     & $10$      & $50000$   & $10000$   \\ \hline
    \multirow{1}{*}{CIFAR-10}       & $32\times 32\times 3$             & $10$      & $50000$   & $10000$   \\ \hline
    \multirow{1}{*}{CIFAR-100}      & $32\times 32\times 3$             & $100$     & $50000$   & $10000$   \\ \hline
    \multirow{1}{*}{SVHN}           & $32\times 32\times 3$             & $10$      & $73257$   & $26032$   \\ \hline
    \multirow{1}{*}{STL-10}         & $96\times 96\times 3$             & $10$      & $5000$    & $8000$    \\ \hline
    \multirow{1}{*}{GTSRB}          & $(15-250)\times(15-250)\times 3$  & $43$      & $39297$   & $12672$   \\  \hline
\end{tabular}
\caption{Different image datasets, and their properties, which are used as a part of the proposed model repository.}
\label{table_data}
\end{table*}

\noindent Similarly, the model catalog has six text datasets. The properties and details of each of these datasets are provided in Table 2.

\begin{table*}[!h]
\centering
\begin{tabular}{|l|l|l|}
    \hline
    Dataset                                 & \#Samples & \#Classes \\ \hline \hline
    \multirow{1}{*}{AG-News}                & $127600$  & $4$       \\ \hline
    \multirow{1}{*}{DBPedia}                & $63000$   & $14$      \\ \hline
    \multirow{1}{*}{Yahoo Answers}          & $146000$  & $10$      \\ \hline
    \multirow{1}{*}{YELP Reviews}           & $700000$  & $5$       \\ \hline
    \multirow{1}{*}{YELP Review Polarity}   & $598000$  & $2$       \\ \hline
    \multirow{1}{*}{SST}                    & $11855$   & $5$       \\ \hline
\end{tabular}
\caption{Different text datasets, and their properties, which are used as a part of the proposed model repository.}
\label{table_data}
\end{table*}

The ensemble of classifiers used over the different features extracted, contains five popular classifiers: (i) Naive Bayes (NB), (ii) Random
Decision Forest (RDF), (iii) Boosted Gradient Trees
(BGT), (iv) Multilayer Perceptron (MLP), and (v) Support Vector Machines (SVM). Table 3 provides different hyperparameters used for each of these classifiers. While some of these hyperparameters are chosen using human expertise and popular defaults, most of the values are obtained through extensive search and experimentation.
\begin{table*}[h]
\centering
\begin{tabular}{|l|l|}
    \hline
    Classifier                  & Parameters \\ \hline
    \hline
    {Multinomial Naive Bayes}   & Nil   \\  \hline
    {Random Decision Forest}    & Depth:25          \\  & NumEstimators:250 \\ \hline
    {Boosted Gradient Trees}    & Depth:25          \\  & NumEstimators:250 \\ \hline
    {Multi Layer Perceptron}    & Solver:LBFGS      \\  & NumHiddenLayers:2 \\  & HiddenSizes: 256,128  \\ \hline
    {Support Vector Machines}   & Kernel: Linear    \\  & \\ \hline
\end{tabular}
\caption{Different features and its parameters used in constructing the ensemble of classifiers model.}
\label{table_class}
\end{table*}

\section{Additional Details: Unsupervised Model Encoder}

To generate valid random text based deep learning models for learning the langauge modeling part, the following grammar is used:
\begin{enumerate}
    \item RNN cell can be RNN/LSTM/GRU
    \item Pooling can be last/max/mean
    \item 300 dimensional embedding, GloVe pretrained
    \item 256 dimensional hidden size
    \item 2 layers, 0.5 dropout between layers
    \item bidirectional learning
    \item Adam optimizer, 0.001 learning rate
    \item gradient clipping at gradient norm of 5
    \item Weight decay of 1e-4
    \item 15 epochs, batch size of 128
\end{enumerate}

\section{Additional Details: Accuracy Prediction}
Hyperparameters for training baseline models and finetuning models is as follows:
\begin{enumerate}
    \item Learning rate 0.1
    \item Weight decay 5e-4
    \item SGD with 0.9 momentum
    \item 128 batchsize, 100 epochs with early stopping
   \item Transformations / Augmentation: RandomCrop (CenterCrop during test), Random Horizontal Flip (except for digit datasets), Mean/Variance Normalization
\end{enumerate}

\section{Implications on Text Classification Datasets and Models}
Having observed the efficacy of our proposed pipeline in a practical application in fine-grained computer vision, we attempted the ambitious goal of trying the same for text classification. We constructed a repository of six datasets: (i) AG-News, (ii) DBPedia, (iii) Yahoo Answers, (iv) YELP Reviews, (v) YELP Review Polarity, and (vi) SST. For the model repository, we have (i) LSTM, (ii) GRU, and (iii) RNN cells, with bidirectional / unidirectional variants, as well as 1-layer / 2-layer variants, for a total of $12$ variants. We observe that the proposed pipeline, as it exists currently, does not excel at recommending text classification models and we derive the following insights from our initial experiments:

\begin{figure*}[t]
	\includegraphics[width=0.9\textwidth]{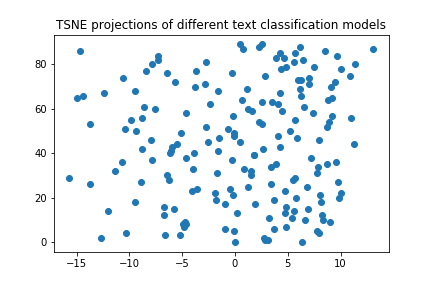}
	\caption{The tSNE representation of $90$ text classification model representations obtained using our unsupervised model encoder.}
	\label{fig:dataset2vec566}
\end{figure*}

\begin{enumerate}
    \item \textbf{Unsupervised Model Encoder:} A total of $19$ unique layers with a vocabulary of $148$ tokens was used to simulate and generate $346,500$ valid text classification models with depth varying from $1$ till $36$. The \textit{word2vec} and the model architecture encoder (explained in section 3.1) were trained from scratch on the generated text classification architectures. However, for the obtained representation we did not observe good clusters in the corresponding tSNE space, as shown in Figure~\ref{fig:dataset2vec566}. This is potentially due to the lack of diversity in the generated RNN based architectures. Upon plotting the tSNE reduced representations of CNN and RNN architectures, we clearly obtained two clusters proving that the unsupervised model encoder learns the representations to some extent.
    \item \textbf{Dataset Similarity:} To find the similarity between two datasets, we use a similar ensemble based technique (as explained in section 3.2). There are four feature extractors used for the text modality: (i) BoW-TF (ii) BoW-TF-IDF (iii) InferSent~\cite{conneau2017supervised} (iv) Skip-thoughts~\cite{kiros2015skip}. Five popular classifiers are used in the ensemble: (i) Naive Bayes (NB), (ii) Random Decision Forest (RDF), (iii) Boosted Gradient Trees (BGT), (iv) Multilayer Perceptron (MLP), and (v) Support Vector Machines (SVM). However, we observed that there was negative correlation between the predicted performance and the actual performance (as compared to Table 1). This could be possible because the similarity between two text datasets could be highly sensitive to the overlapping vocabulary space, unlike in images where there exists some abstract overlapping concepts such as edges, corners, and basic shapes.
\end{enumerate}

\end{document}